\documentclass[letterpaper, 10 pt, conference]{ieeeconf}  
\usepackage{amssymb}
\usepackage{amsmath}
\usepackage{graphicx}
\usepackage{graphics}
\usepackage{blkarray}
\usepackage{booktabs}
\usepackage{xcolor} 
\usepackage{bbold}
\usepackage{caption}
\usepackage{subcaption}
\usepackage{multirow}
\usepackage{comment}
\usepackage{epstopdf}
\usepackage{algorithm2e}
\usepackage{tikz}
\usepackage{todonotes}
\usetikzlibrary{arrows,matrix,positioning}
\graphicspath{{./figures/}}
\usepackage{caption}
\usepackage[labelformat=simple]{subcaption}
\usepackage{soul}
\usepackage{MnSymbol}
\usepackage{cite}

\DeclareCaptionLabelSeparator{periodspace}{.\quad}
\IEEEoverridecommandlockouts                              
\newtheorem{lemma}{Lemma}
\newtheorem{remark}{Remark}
\newtheorem{assumption}{Assumption}
\newtheorem{example}{Example}

\newtheorem{corollary}{Corollary}
\newtheorem{problem}{Problem} 

\newtheorem{definition}{Definition}

\makeatletter
\newcommand*{\rom}[1]{\expandafter\@slowromancap\romannumeral #1@}
\makeatother

\newcommand{\qed}{\nobreak \ifvmode \relax \else
      \ifdim\lastskip<1.5em \hskip-\lastskip
      \hskip1.5em plus0em minus0.5em \fi \nobreak
      \vrule height0.75em width0.5em depth0.25em\fi}

\mathchardef\mhyphen="2D

\newcommand{\bt}{\mathcal{T}}

\title{\LARGE \bf
Adaptive Fault Tolerant Execution of \\  Multi-Robot Missions using Behavior Trees}

 \author{Michele Colledanchise,$^{\dagger}$ Alejandro Marzinotto,$^{\dagger}$ Dimos V. Dimarogonas,$^{\ddagger}$ and Petter \"Ogren$^{\dagger}$  
\thanks{$^{\dagger}$The author is with the Centre for Autonomous Systems, Computer Vision and Active Perception Lab, School of Computer Science and Communication, The Royal Institute of Technology - KTH, Stockholm, Sweden.}
\thanks{$^{\ddagger}$The author is with the Centre for Autonomous Systems, Automatic Control Lab, School of Electrical Engineering, The Royal Institute of Technology - KTH, Stockholm, Sweden.}%
}

\begin{document}
\maketitle
\thispagestyle{empty}
\pagestyle{empty}
\begin{abstract}

Multi-robot teams offer possibilities of improved performance and fault tolerance, compared to single robot solutions.
In this paper, we show how to realize those possibilities 
when starting from a single robot system controlled by a Behavior Tree (BT).
By extending the \emph{single robot BT} to a \emph{multi-robot BT}, we are able to combine the fault tolerant properties of the BT, in terms of built-in fallbacks, with the fault tolerance inherent in multi-robot approaches, in terms of a faulty robot being replaced by another one. Furthermore, we improve performance by identifying and taking advantage of the opportunities of parallel task execution, that are present in the single robot BT. 
Analyzing the proposed approach,
we
present results regarding how mission performance is affected by \emph{minor faults} (a robot losing one capability) as well as \emph{major faults} (a robot losing all its capabilities). Finally, a detailed example is provided to illustrate the approach.

\end{abstract}

%
\section{Introduction}
\label{IN}

\begin{figure}[t]
\centering
\includegraphics[width=0.8\columnwidth]{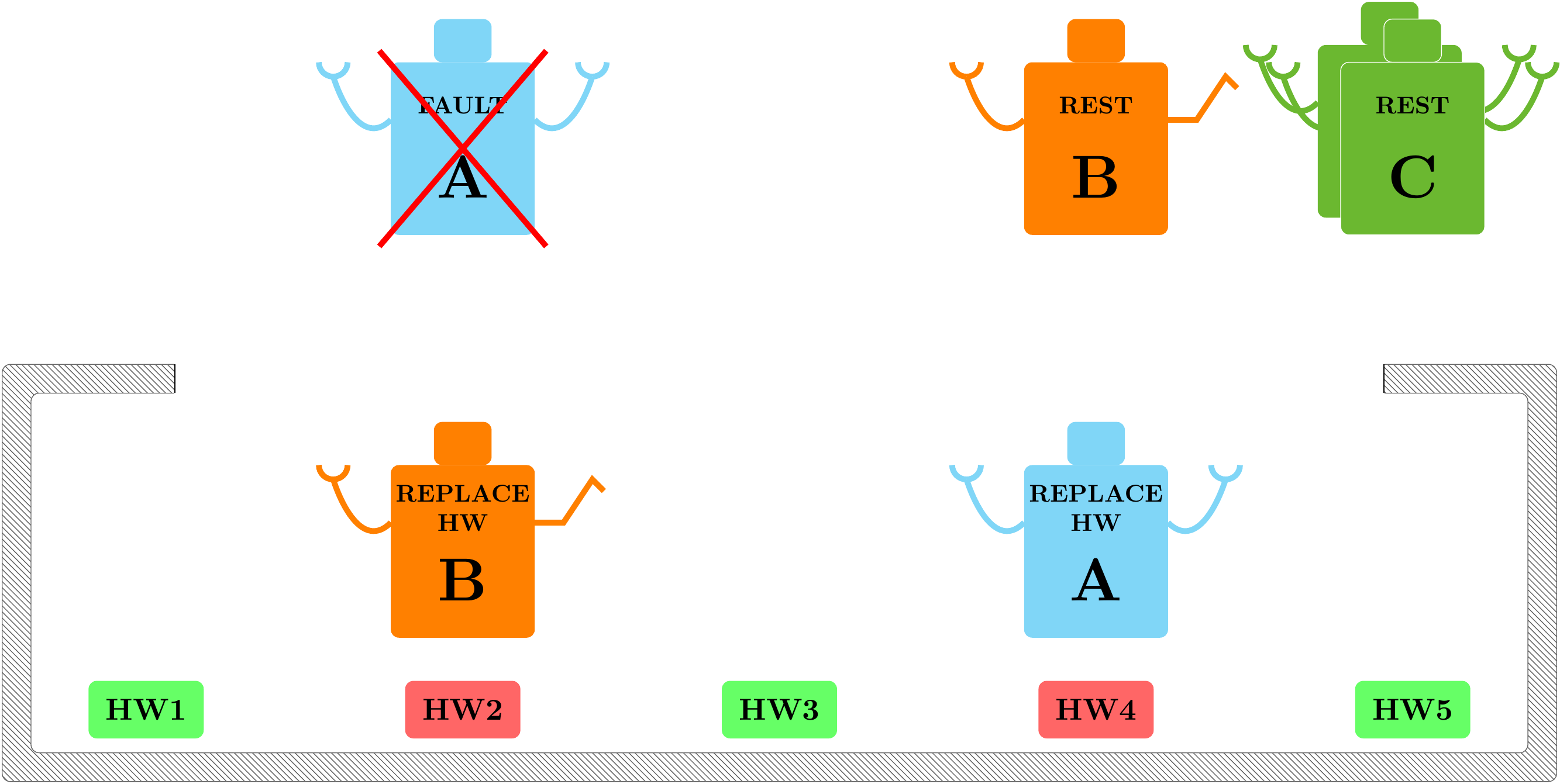}
\caption{Snapshot of a multi-robot mission execution. Two robots are replacing the broken parts, HW2 and HW4, (bottom) of a machine. One robot is out of order (top left), and tree robots are resting (top right).}
\label{IN.fig.front}
\end{figure}

Imagine a robot that is designed to perform maintenance on a given machine. The robot can open and close the cover, do fault detection and replace broken hardware (HW) components. However, this robot is fairly complex and easily breaks down. Thus, it is desirable to replace this big versatile robot with a team of smaller specialized robots, as shown in Figure~\ref{IN.fig.front}.
 This paper shows how to modify the single robot Behavior Tree (BT) controlling the original robot, to a multi-robot BT, to be run in parallel on all new robots, thereby improving both  fault tolerance and performance of the original BT.
 
 The fault tolerance of the single robot BT, in terms of \emph{fallbacks}, is improved by also adding
   the tolerance afforded by the redundancy achieved through having multiple robots, and the performance of the single robot BT is improved by executing the right tasks in parallel.
  In detail, the BT includes both so-called \emph{sequence} and \emph{fallback} compositions, as explained below.
 Some tasks that were earlier done in sequence, such as diagnosing different HW components, are now done in parallel, while other tasks that were also done in sequence, such as opening the cover and then diagnosing HW, are still done in a sequence.
 Similarly, some tasks that were earlier done as fallbacks, such as searching in storage 2 if no tools were found in storage 1, are now done in parallel, while other fallbacks, such as using the small screwdriver if the large one does not fit, are not done in parallel. Thus fault tolerance as well as performance is improved.
 
 The contribution of this paper is that we show how fault tolerance and performance can be improved in a single robot BT, by adding more robots to the team and extending the BT into a multi-robot BT, using two important modifications. First we add a task assignment functionality, and then we identify and adapt the sequences and fall back compositions of the BT that are possible to execute in parallel.
 
The outline of this paper is as follows. 
First, in Section~\ref{RW} we review related work. Then, in Section~\ref{BG} we give an overview of the classical formulation of BTs. The main problem is stated in  Section~\ref{PF} and the proposed solution is given in Section~\ref{PS}. An extensive example illustrating the proposed approach is presented in Section~\ref{ME}, and  we conclude the paper in Section~\ref{Conc}.

\section{Related Work}
\label{RW}
BTs are a recent alternative to Controlled Hybrid Systems (CHSs) for reactive fault tolerant execution of robot tasks, and they were first introduced in the computer gaming industry
\cite{isla2005handling, champandard2007understanding, isla2008halo}, to meet their needs of modularity, flexibility and reusability in artificial intelligence for in-game opponents.
 Their  popularity lies in its ease of understanding, its recursive structure, and its usability, creating a growing attention in academia
\cite{lim2010evolving, perez2011evolving, shoulson2011parameterizing, bojic2011extending, ogren, Bagnell2012b, klockner2013,Colledanchise14-2}.
In most cases, CHSs have memoryless transitions, i.e. there is no information where the transition took place from, a so-called \emph{one way control transfer}. In BTs the equivalent of state transitions are governed by calls and return values being sent by parent/children in the tree structure, this information passing is called a \emph{two way control transfers}. In programming languages, the replacement of a one way (e.g., GOTO statement) with a two way control transfer (i.e., Function Calls) made an improvement in readability and reusability \cite{Dijkstra:1968:LEG:362929.362947}.
Thus, BTs exhibit similar advantages as gained moving from GOTO to Function Calls in programming  in the 1980s. Note however, 
we do \emph{not} claim that BTs are better than CHSs from a purely theoretical standpoint.
Instead,  the main advantage of using BTs lies is in its ease of use, in terms of modularity and reusability, \cite{ogren}.
BTs were first used in \cite{isla2005handling,champandard2007understanding}, in high profile computer games, such as the HALO series. 
Later work merged machine learning techniques with BTs' logic \cite{lim2010evolving,perez2011evolving}, making them more flexible in terms of parameter passing \cite{shoulson2011parameterizing}.
The advantage of BTs as compared to a general Finite State Machines (FSMs) was also the reason for extending the JADE agent Behavior Model with BTs in \cite{bojic2011extending}, and  the benefits of using BTs to control complex missions of Unmanned Areal Vehicles (UAVs) was described in \cite{ogren}.
In \cite{klockner2013} the modular structure of BTs addressed the formal verification of mission plans.

In this work, we show how a plan for a multi-robot system can be addressed in a BT fashion, gaining all the advantages aforementioned in addition to  the scalability that distinguishes a general  multiagent system.  Many existing works~\cite{Chu11,Karaman08,Ulusoy12} stressed the problem of defining local tasks to achieve a global specification emphasizing the advantages of having a team of robots working towards a global goal. Moreover, \cite{Castelfranchi98towardsa,Falcone01the} introduce the concept of task delegation among agents in a multi-agent system, dividing task specification in \emph{closed delegation}, where goal and plan are predefined, and \emph{open delegation} where either only the goal is specified while the plan can be chosen by the agent, or the specified plan describes abstractly what actions have to be taken, giving to the agent some freedom in terms of how to perform the delegated task. In \cite{Doherty12}, it is shown how verifying the truth of preconditions on single agents becomes equivalent to checking the fulfillment of a global robot network through recursive calls, using a tree structure called \emph{Task Specification Trees}.

Recent works present some advantages of implementing BTs in robotics applications~\cite{Marzinotto14,ogren} making comparisons with the CHSs, highlighting their modularity and reusability. In \cite{Bagnell2012b}, BTs were used to perform autonomous robotic grasping. In particular, it was shown how BTs enabled the easy composition of primitive actions into complex and robust manipulation programs.
Finally, in~\cite{Colledanchise14} performance measures, such as success/failure probabilities and execution times, were estimated for plans using BTs. However, BTs are mostly used to design single agent behavior and, to the best of our knowledge, there is no rigorous framework in  academia using the classical formulation of BTs for multi-robot systems.  


\section{Background: Formulation of BTs}
\label{BG}
\begin{figure}[h]
\centering
\includegraphics[width=0.6\columnwidth]{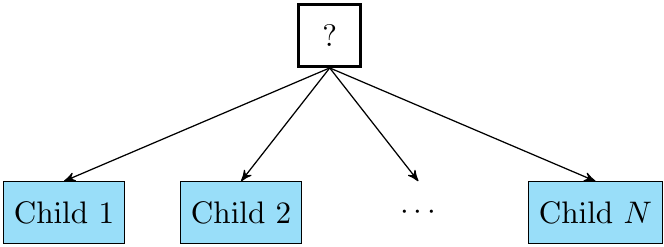}
\caption{Graphical representation of a selector node with $N$ children.}
\label{bg.fig.sel}
\end{figure}

\begin{figure}[h]
\centering
\includegraphics[width=0.6\columnwidth]{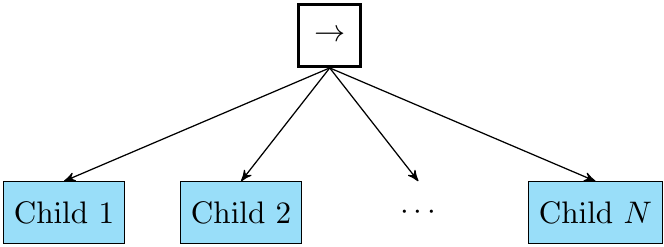}
\caption{Graphical representation of a sequence node with $N$ children.}
\label{bg.fig.seq}
\end{figure}

\begin{figure}[h]
\centering
\includegraphics[width=0.6\columnwidth]{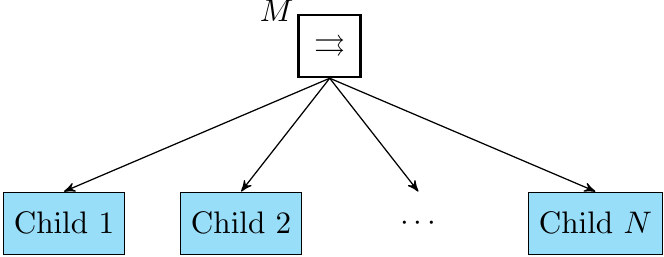}
\caption{Graphical representation of a parallel node with $N$ children.}
\label{bg.fig.par}
\end{figure}

Here, we briefly introduce an overview of BTs, the reader can find a detailed description in \cite{ogren}.\\
A BT is defined as a directed rooted tree where nodes are grouped into  control flow nodes,  execution nodes, and a root node. In a pair of connected nodes we call  the outgoing node \emph{parent} and  the incoming node \emph{child}. 
Then, the root node has no parents and only one child, 
 the control flow nodes have one parent and at least one child, and
 the execution nodes are the leaves of the tree (i.e. they have no children and one parent).
 Graphically, the children of a control flow node are sorted from its bottom left to its bottom right, as depicted in Figures~\ref{bg.fig.sel}-\ref{bg.fig.par}.
The execution of a BT starts from the root node. It sends \emph{ticks}~\footnote{A tick is a signal that enables the execution of a child.} to its child. When a generic node in a BT receives a tick from its parent,  its execution starts and it returns to its parent a status \emph{running} if its execution has not finished yet, \emph{success} if its execution is accomplished (i.e. the execution ends without failures), or \emph{failure} otherwise.\\ 
Here we draw a distinction between three types of control flow nodes (selector, sequence, and parallel) and between two types of execution nodes (action and condition). Their execution is explained below.
\paragraph*{Selector (also known as Fallback)}
When the execution of a selector node starts (i.e. the node receives a tick from its parent), then the node's children are executed in succession from left to right, 
until a child returning success or running is found. Then this message is returned to the parent of the selector.
 It returns failure only when all the children return a status failure.
The purpose of the selector node is to robustly carry out a task that can be performed using several different approaches (e.g. a motion tracking task can be made using either a 3D camera or a 2D camera) by performing each of them in succession until one succeeds.
The graphical representation of a selector node is a box with a ``?", as in Fig.~\ref{bg.fig.sel}. \\ A finite number of BTs $\bt_1,\bt_2,\ldots,\bt_N$ can be composed into a more complex BT, with them as children, using the selector composition: $\bt_0=\mbox{Selector}(\bt_1,\bt_2,\ldots,\bt_N)$.
\paragraph*{Sequence}
When the execution of a sequence node starts, then the node's children are executed in succession from left to right, returning to its parent a status failure (running) as soon as the a child that returns failure (running) is found. It returns success only when all the children return success.
The purpose of the sequence node is to carry out the tasks that are defined by a strict sequence of sub-tasks, in which all have to succeed (e.g. a mobile robot that has to move to a region ``A" and then to a region ``B"). 
The graphical representation of a sequence node is a box with a ``$\rightarrow$", as in Fig.~\ref{bg.fig.seq}.\\ A finite number of BTs $\bt_1,\bt_2,\ldots,\bt_N$ can be composed into a more complex BT, with them as children, using the sequence composition: $\bt_0=\mbox{Sequence}(\bt_1,\bt_2,\ldots,\bt_N)$.
\paragraph*{Parallel} 
When the execution of a parallel node starts, then the node's children are executed in succession from left to right without waiting for a return status from any child before ticking the next one. It returns success if a given number of children $M\in \mathbb{N}$ return success, it returns failure when the children that return running and success are not enough to reach the given number, even if they would all   return success. It returns running otherwise. The purpose of the parallel node is to model those tasks separable in independent sub-tasks performing non conflicting actions (e.g. a multi object tracking can be performed using several cameras).
The parallel node is graphically represented by a box with ``$\rightrightarrows$" with the number $M$ on top left, as in Fig.~\ref{bg.fig.par}. 
A finite number of BTs $\bt_1,\bt_2,\ldots,\bt_N$ can be composed into a more complex BT, with them as children, using the parallel composition: $\bt_0=\mbox{Parallel}(\bt_1,\bt_2,\ldots,\bt_N,M)$.
\begin{figure}[h]
        \centering
       ~ 
        \begin{subfigure}[b]{0.3\columnwidth}
                \centering
                \includegraphics[width=0.3\columnwidth]{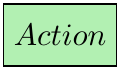}
                \caption{Action node.}
                \label{bg.fig.act}              
        \end{subfigure}
        ~ 
        \begin{subfigure}[b]{0.3\columnwidth}
                \centering
                \includegraphics[width=0.5\columnwidth]{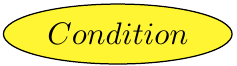}
                \caption{Condition node.}
                \label{bg.fig.cond}
        \end{subfigure}
        \caption{Graphical representation of action and condition nodes.}
\end{figure}
\paragraph*{Action}
When an action node starts its execution, then it returns
 success if the action is completed and failure if the action cannot be completed. Otherwise it returns
 running. The action node is represented in Fig.~\ref{bg.fig.act} 
\paragraph*{Condition}
The condition node checks if a condition is satisfied or not. The return status is success or failure accordingly and it is never running. The condition node is represented in Fig.~\ref{bg.fig.cond}. 
\paragraph*{Root}
The root node is the node that generates ticks. It is graphically represented by a box labeled with ``$\varnothing$''.
\section{Problem Formulation}
\label{PF}
The problem considered in this work is to define identical decentralized  local BT controllers for the single robots of a multi-robot system to achieve a global goal.
We first give some assumptions used throughout the paper, we then state the main problem.\\
Let $\mathcal{S}$ be a set of symbols describing atomic actions (e.g. perform grasp, go to position) and let $\mathcal{R}\in \mathbb{N}$ be the set of robots in a multi-robot system, each of which can perform a finite collection of local tasks $\mathcal{L}_i\subseteq \mathcal{S}$, $i\in\mathcal{R}$ and $n=|\mathcal{R}| $ and let $\bar{\mathcal{L}} = \bigcup_{i\in\mathcal{R}} \mathcal{L}_i$ be the set of all the local tasks. Now let $\mathcal{G}=\{g_1,g_2,\ldots,g_{v}\}$ be a finite collection of global tasks executable by the multi-robot system and $P_k$ be the set of global tasks running in parallel with $g_k$. 
We define $\psi(g_k)$ as the finite set of local tasks which have to return success for  the global task $g_k$ to succeed. \\
Finally, let  $\nu: \mathcal{G} \times \bar{\mathcal{L}} \to \mathbb{N}$ and $\mu: \mathcal{G} \times \bar{\mathcal{L}} \to \mathbb{N}$  respectively be functions that give the minimum number of robots needed, and the maximum number of robots assignable, to perform a local task, in order to accomplish a global task.
Given these, we state the following assumptions:
\begin{assumption}
The minimum number of robots required to execute global tasks running in parallel does not exceed the total number of robots in the system: $\sum_{g_k \in P_h}\sum_{l_j \in \psi(g_k)}\nu(g_k,l_j) \leq n \; \forall g_h\in\mathcal{G} $.
\label{PF:ass:number}
\end{assumption}
\begin{assumption}
Each global task in $\mathcal{G}$ can be performed by assigning to some robots in $\mathcal{R}$  one of their local tasks: $\psi(g_k)~\subseteq~2^{\bar{\mathcal{L}}}$ $\forall g_k \in \mathcal{G}$.
\label{PF:ass:globalocal}
\end{assumption}
\begin{problem}
Given a multi-robot system defined as above and a global goal that satisfies Assumptions~\ref{PF:ass:number}-\ref{PF:ass:globalocal}, design a local BT controller for each single robot to achieve this goal.
\label{PF:problem}
\end{problem}
\section{Proposed Solution} 
\label{PS}
In this section, we will first 
give an informal description of the proposed solution, involving the three subtrees $\bt_G$ (global tasks), $\bt_A$ (task assignment) and $\bt_{L_i}$(local tasks). Then, we describe the design of these subtrees in detail. To illustrate the approach we give a brief example, followed by a description of how to create a multi agent BT from a single agent BT. Finally, we analyze the fault tolerance of the proposed approach.

The multi agent BT is composed of three subtrees running in parallel, returning success only if all three subtrees return success. 
\begin{equation}
\bt_i=\mbox{Parallel}(\bt_A,\bt_G,\bt_{Li},3)
\label{PS.eq.tree}
\end{equation}

\begin{remark}
Note that all robots run identical copies of this tree, including computing assignments of all robots in the team, but each robot only executes the task that they themselves are assigned to, according to their own computation.
\end{remark}

The first subtree, $\bt_A$ is doing task assignment. Parts of the performance and fault tolerance of the proposed approach comes from this feature, making sure that a broken robot is replaced and that robots are assigned to the tasks they do best.

The second subtree,  $\bt_G$ takes care of the overall mission progression. 
This tree includes information on  task that needs to be done in sequence (using sequence composition), tasks that could either be done in sequence or in parallel (using parallel $M=N$ compositions), fallback tasks that should be tried in sequence (using selector compositions), and fallbacks that can be tried in parallel (using parallel $M=1$ compositions). The execution of $\bt_G$ provides information to the assignment in $\bt_A$. The fault tolerance of the proposed approach is improved by  the fallbacks encoded in $\bt_G$, and the performance is improved by parallelizing actions whenever possible, as described above.

The third subtree, $\bt_{L_i}$ uses output from the task assignment to actually execute the proper actions. We will now define the three subtrees in more detail.

%
%
%
%
%
%

\begin{figure}[h]
\centering
\begin{subfigure}[b]{0.4\columnwidth}
\centering
\includegraphics[width=\columnwidth]{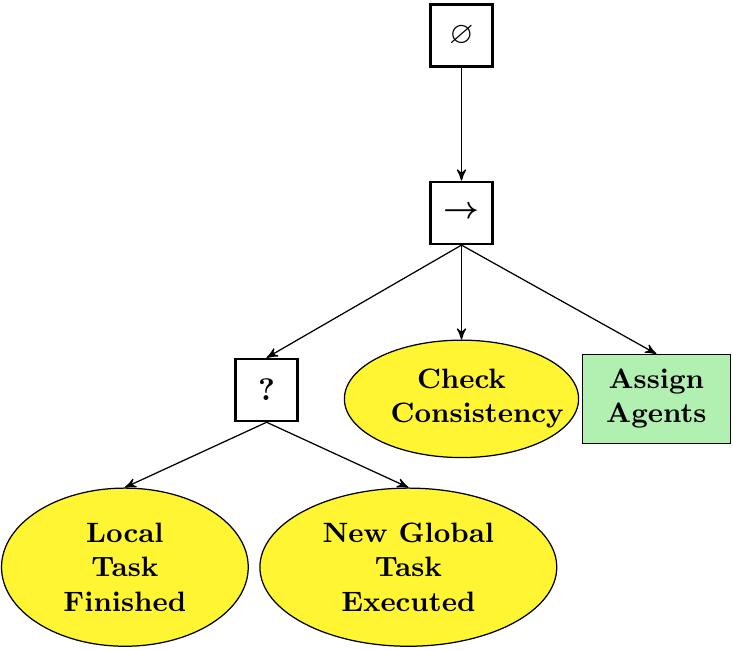}
\caption{Task Assignment tree }
\label{PS.fig.ta}
\end{subfigure}
\;
\begin{subfigure}[b]{0.4\columnwidth}
\centering
\includegraphics[width=0.9\columnwidth]{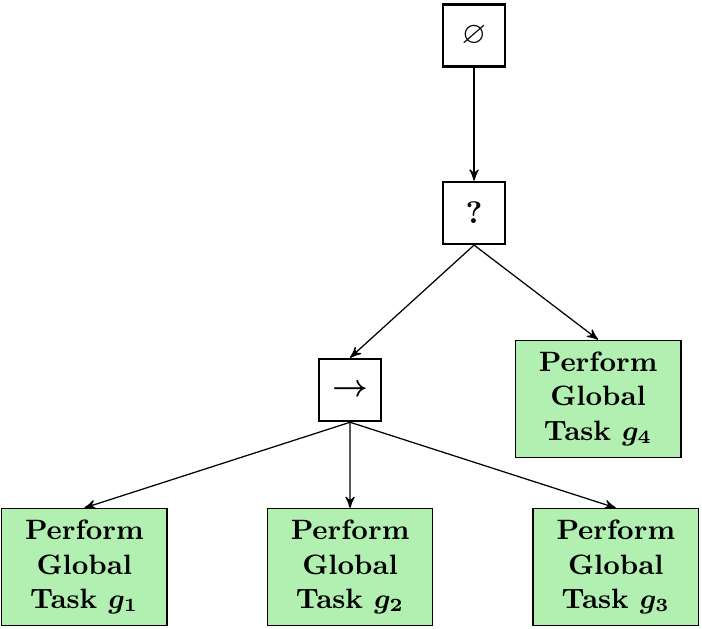}
\caption{Global tasks tree}
\label{PS.fig.tg}
\end{subfigure}
\caption{Example of BT modeling a global task}\label{PS.fig.tatg}
\end{figure}

\paragraph{Definition of $\bt_A$}
This tree is responsible for the reactive optimal task assignment, deploying robots to different local tasks according to the required scenario. The reactiveness lies in changes of constraints' parameters in the optimization problem, depicting the fact that the number of robots needed to perform a given task changes during its execution.\\ The assignment problem is solved using any other method used in optimization problem. Here we suggest the following, however the user can replace the solver without changing the BT structure: 
\begin{equation}
\begin{aligned}
& \underset{}{\text{minimize}}
& & \mhyphen \sum_{i\in\mathcal{R}}\sum_{l_j \in \bar{\mathcal{L}}}p(i,l_j )r(i,l_j ) \\
& \text{subject to:}
& & a_{j} \leq \sum_{i\in\mathcal{R}} r(i,l_j )\leq b_{j} \; \forall  l_j\in {\mathcal{L}_i} \\
& & & \; \; \;  \; \;\; \;  \sum_{l_j\in \bar{\mathcal{L}}} r(i,l_j)\leq 1 \; \forall i\in\mathcal{R} \\
& & &  r(i,j)\in \{0,1\}
\end{aligned}
\label{PS.eq.ass}
\end{equation}
where $r: \mathcal{R} \times \bar{\mathcal{L}} \to \{0,1\}$ is a function that represents the assignment of robot $i$ to a local task $l_j$, taking value $1$ if the assignment is done and $0$ otherwise; $p: \mathcal{R} \times \bar{\mathcal{L}} \to \bar{\mathbb{R}}$ is a function assigning a performance to robot $i$ at executing the task $l_j$ (if the robot $i$ cannot perform the task $l_j$ i.e., $l_j \notin \mathcal{L}_i$ then $p(i,l_j)=\mhyphen \infty$). ${a}_{j}, b_{j} \in \mathbb{N}$ are respectively the minimum and the maximum number of robots assignable to the task $l_j$ and they change during the execution of~\eqref{PS.eq.tree} and ${a}_j\in \mathbb{N}$  is set to a positive value upon requests from $\bt_G$. At the beginning they are initialized to $0$, since no assignment is needed. When the execution of a global task $g_k$ starts,  $a_{j}$ and $b_{j}$ are set respectively to $\nu(g_k,l_j)$ and $\mu(g_k,l_j)$ $\forall l_j \in \psi(g_k)$. Then when a local task $l_j$ finishes its execution (i.e., it returns either success or failure),  $a_{j}$ and $b_{j}$ decrease by $1$ until $a_{j}=0$, since the robot executing $l_j$ can be assigned to another task while the other robots are still executing their local task. When the task $g_k$ finishes, both $a_{j}$ and $b_{j}$ are set back to zero.\\
The tree $\bt_A$ is shown in Fig.~\ref{PS.fig.ta}. The condition ``Local Task Finished" returns success if a robot has succeeded or failed a local task, failure otherwise. For ease of implementation a robot can communicate to all the other when has finished a local task. The condition ``New Global Task Executed" returns success if it is satisfied, failure otherwise.\\ The condition ``Check Consistency" checks if  the constraints of the optimization problem are consistent with each other, since such constraints change their parameters during the tree execution. The addition of a constraint in~\eqref{PS.eq.ass} effects the system only if a solution exists. This condition is crucial to run a number of global tasks in parallel according to the number of available robots (see Example~\ref{PS.ex1} below). \\ The action ``Assign Agents" is responsible for the task assignment, it returns running while the assignment routine is executing, it returns success when the assignment problem is solved and it returns failure if the optimal value is $\infty$.

\paragraph{Definition of $\bt_G$}
This tree is designed to achieve the global goal defined in Problem~\ref{PF:problem} (i.e., the global goal is achieved if the tree returns success) executing a set of global tasks.\\ 
An example of $\bt_G$ is depicted in Fig.~\ref{PS.fig.tg}.
The execution of the action ``Perform Global Task $g_k$" requires that some robots have to be assigned, then it sets  $b_{j}=\mu(g_k,l_j) \; \forall l_j \in \psi(g_k)$. In the tree $\bt_A$ the condition ``New Global Task Executed" is now satisfied, making a new assignment if the constraints are consistent with each other. Finally action ``Perform Global Task $g_k$" returns success if the number of local tasks $l_j \in \psi(g_k)$ that return success is greater than $\nu(g_k,l_j)$ (i.e the minimum amount of robots needed have succeeded) it returns failure if the number of local tasks in $\psi(g_k)$ that return failure is greater than $\sum_{i\in \mathcal{R}} r(i,j)-\nu(g_k,l_j)$ (i.e., some robots have failed to perform the task, the remaining ones are not enough to succeed), it returns running otherwise. The case of translating a single robot BT to a multi robot BT is described in Section \ref{SBTMBT} below.

\paragraph{Definition of $\bt_{Li}$}
This tree comprises the planner of a single agent. Basically it executes one of its subtrees upon request from the $\bt_A$, the request is made by executing the action ``Assign Agents".\\ 

\begin{figure}[h]
\centering
\includegraphics[width=\columnwidth]{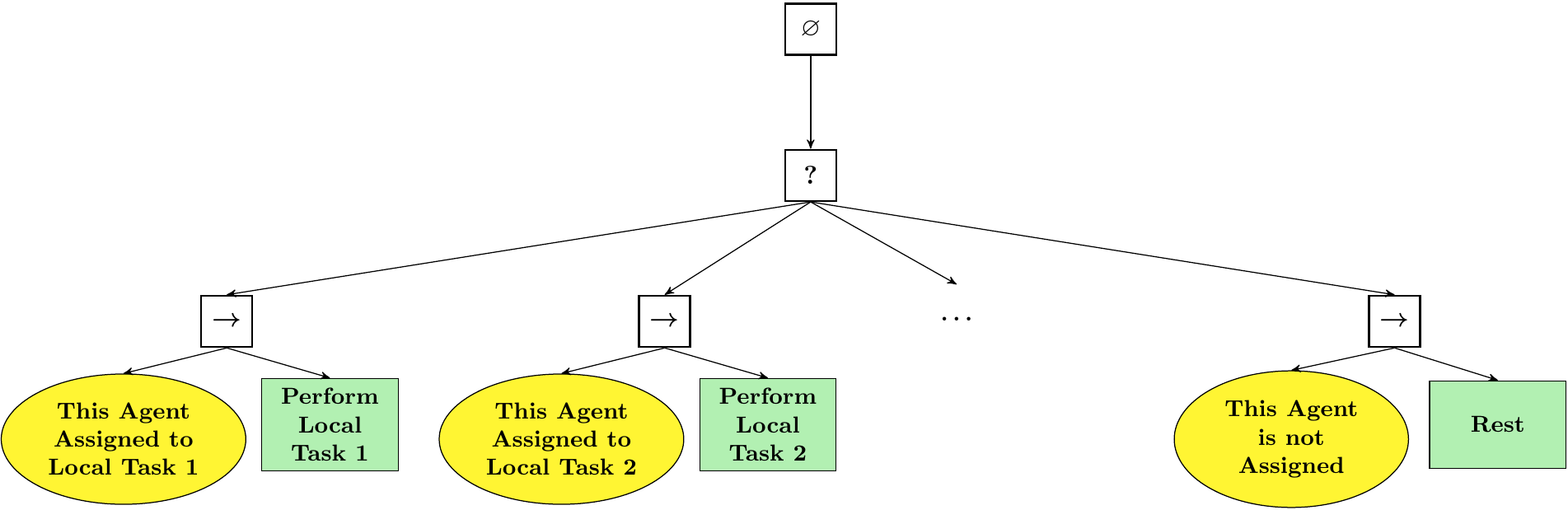}
 \caption{Example of BT modeling a local task on a single agent}\label{multi.fig.localex}
\end{figure}
\begin{example}
By way of example, we consider a system that is to explore $2$ different areas. Intuitively, different areas can be explored by different robots at the same time but if only a single robot is available those areas can only be explored in a sequence. The local trees $\bt_{L_i}$ are as depicted in Fig.~\ref{PS.fig.lex1} and the global task tree $\bt_G$ as in Fig.~\ref{PS.fig.gex1}. \\ Considering first the case in which there is only one robot available. When the execution of \emph{Explore Area $A$} in $\bt_G$ starts, in~\eqref{PS.eq.ass} the constraint that one robot has to be assigned to explore area $A$ is added (i.e. the related $a_j$ and $b_j$ are both set to $1$). The optimization problem is feasible hence the assignment takes place. When the execution of \emph{Explore Area $B$} starts, the related constraint is not consistent with the other constraints, since the only available robot cannot be assigned to two tasks, hence the assignment does not take place. \\ Considering now the case where a new robot is introduced in the system, it is possible to assign two robots. Both  constraints: the one related to \emph{Explore Area $A$} and the one related to \emph{Explore Area $B$}, can be introduced in the optimization problem without jeopardizing its consistency. Those two tasks can be executed in parallel. \\ Note that the two different executions (i.e. \emph{Explore Area $A$} first, then \emph{Explore Area $B$};  \emph{Explore Area $A$} and \emph{Explore Area $B$} in parallel) depend only on the number of available robots, the designed tree does not change.
\begin{figure}[h]
\centering
\begin{subfigure}[b]{\columnwidth}
\centering
\includegraphics[width=0.2\columnwidth]{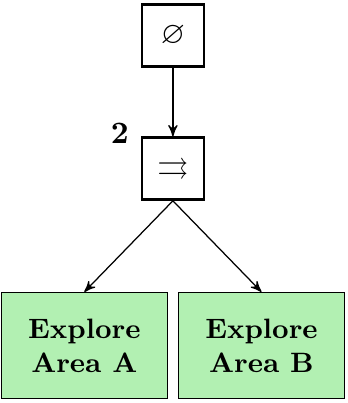}
\caption{$\mathcal{T}_G$ of example~\ref{PS.ex1}}
\label{PS.fig.gex1}
\end{subfigure}
\begin{subfigure}[b]{\columnwidth}
\centering
\includegraphics[width=0.8\columnwidth]{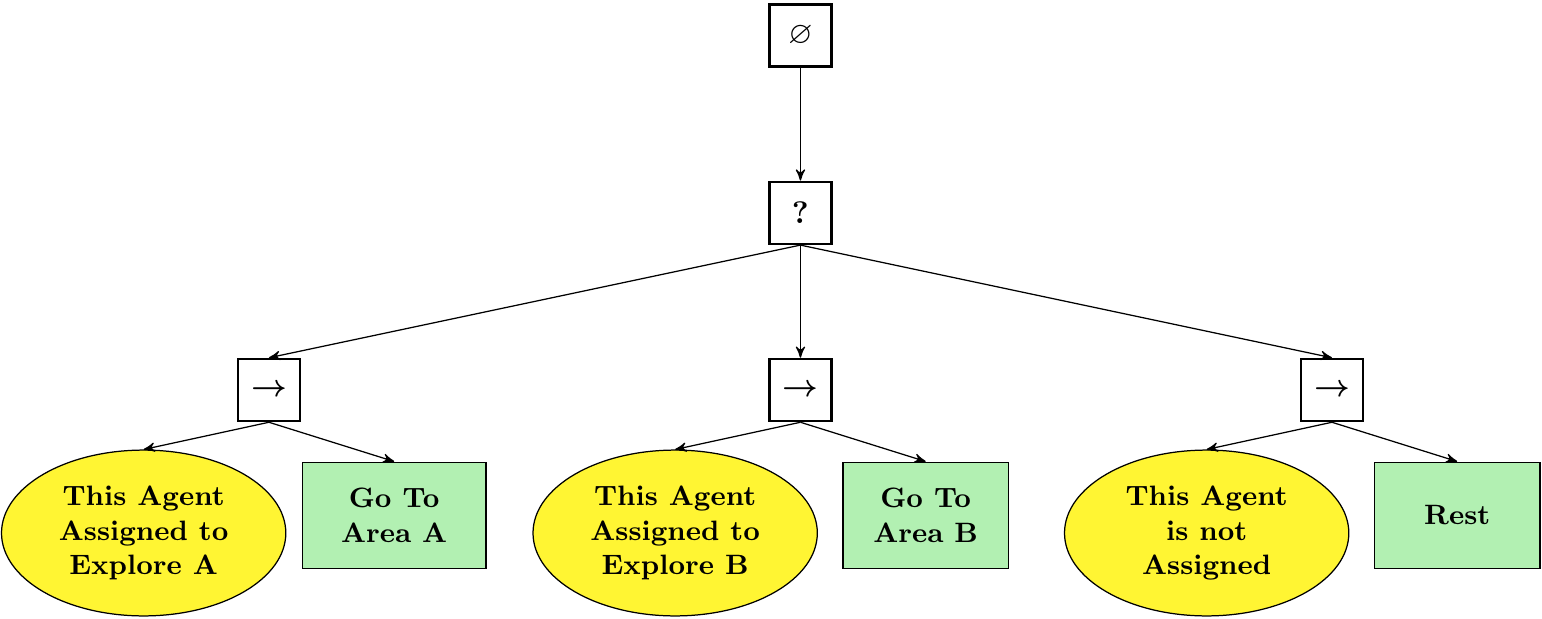}
\caption{$\mathcal{T}_{Li}$ of example~\ref{PS.ex1}}
\label{PS.fig.lex1}
\end{subfigure}
\caption{Trees of example~\ref{PS.ex1}}

\end{figure}
\label{PS.ex1}
\end{example}

\subsection{From single robot BT to multi-robots BT}
\label{SBTMBT}
Here we present how one can design a BT for a multi-robot system reusing the BT designed to control a single robot. Let $\bt_S$ be the tree designed for the single robot execution with $\mathcal{A}_S=\{\alpha_1,\alpha_2,\cdots\}$ being the set of  action nodes, each robot in the multi-robot system will have as $\bt_G$ a tree with the same structure of $\bt_S$ changing the meaning of the action nodes and replacing the control flow nodes with  parallel nodes where possible.\\ The action nodes of $\bt_G$ are the global tasks $\mathcal{G}$ of the tree (i.e. tasks that the entire system has to perform). Their execution has the same effect on the multi-robot system as the effect of the actions $\mathcal{A}_S$ in $\bt_S$. The execution of tasks in $\mathcal{G}$ adds constraints in the optimization problem of the tree $\bt_A$ defined by the sets $\psi(g_k)$, $\forall g_k\in \mathcal{G}$. Those sets, in this case, have all cardinality $1$ (i.e., only one robot is assigned to perform a task in $\mathcal{G}$) since it is a translation from a single robot execution to a multi-robot execution.\\The main advantage of having a team of robots lies in fault tolerance and the possibility of
 parallel execution of some tasks. In the tree $\bt_G$ the user can replace sequence and selector nodes with parallel nodes and $M=N$ and $M=1$ respectively (i.e., if the children of a selector node can be executed in parallel, the correspondent parallel node returns success as soon as one child returns success; on the other hand  if the $N$ children of a sequence node can be executed in parallel, the correspondent parallel node returns success only when all of them return success). The trees $\bt_{Li}$ have the same structure of the one defined in section~\ref{PS}, defining as ``Perform Local Task $k$" the action $\alpha_k$.\\ Then to construct $\bt_G$, first each action node of $\bt_S$ is replaced by a global task $g_k$ with $\psi(g_k)=\alpha_k$ and then the control flow nodes are replaced with parallel nodes where possible. The tree $\bt_A$ is always the one defined in section~\ref{PS}.\\Finally, each robot~$i$ in a multi-robot system runs the tree $\bt_i$ defined in~\eqref{PS.eq.tree}.\\ Here we observe that if the multi-robot system is heterogeneous the trees $\bt_{Li}$ are such that they hold the condition $\bar{\mathcal{L}}\subseteq\mathcal{A}_S$ (i.e. the multi-robot system can perform the tasks performed by the original single robot).

\subsection{Fault Tolerance}
The fault tolerance capabilities of a multi-robot system are due to the redundancy of tasks executable by different robots. In particular, a local task $l_j$ can be executed by every robot $i\in \mathcal{R}$ satisfying the condition $l_j \in \mathcal{L}_i$. Here we make the distinction between \emph{minor} faults and \emph{major} faults of a robot according to their severity. A minor fault involves a single local task $l_j$, i.e. the robot is no longer capable to perform the task $l_j$, here the \emph{level} of a minor fault is the number of local tasks involved. A major fault of a robot implies that the robot can no longer perform any of its local task.
\begin{remark}
A minor fault of level $\lambda$ is different from $\lambda$ minor faults, since the latter might involve different robots.
\end{remark}
\begin{remark}
A major fault involving a robot $i\in \mathcal{R}$ is equivalent to a minor fault of level $|\mathcal{L}_i|$ involving every task in $\mathcal{L}_i$.
\end{remark}
\begin{definition}
A multi-robot system is said to be \emph{weakly fault tolerant} if it can tolerate any minor fault.
\end{definition}
\begin{definition}
A multi-robot system is said to be \emph{strongly fault tolerant} if it can tolerate any major fault.
\end{definition}
\begin{lemma}
A multi-robot system is weakly fault tolerant if and only if for each robot $i\in \mathcal{R}$ and for each local task $l_j\ \in \mathcal{L}_i$ there exists another robot $h\in \mathcal{R}$ such that $l_j \in \mathcal{L}_i \cap \mathcal{L}_h$ and the problem \eqref{PS.eq.ass} is consistent under the constraint $r(i,j) \cdot r(h,j)=0$.

\begin{proof}(\emph{if})
When $r(i,j) \cdot r(h,j)=0$ either robot $i$ or $h$ is not deployed. Assume that the robot $i$ is involved in a minor fault, related to the local tasks $l_j$. Since $l_j \in \mathcal{L}_i \cap \mathcal{L}_h$, $l_j \in \mathcal{L}_h$ then the robot $h$ can perform the local task $l_j$ in place of robot $i$, since it is not deployed.

(\emph{only if}) if the system can tolerate a minor fault then for each robot $i$ there exists another robot $h$ such that the local task related to the fault can be performed by both of them since one took over the other. Hence $l_j \in \mathcal{L}_i$ and $l_j \in \mathcal{L}_h$, this implies that $l_j \in \mathcal{L}_i \cap \mathcal{L}_h$ holds. Moreover, if the fault can be tolerated it means that when $l_j$ is performed either the robot $i$ or robot $h$ is not deployed otherwise the reassignment would not be possible. Hence the condition $r(i,j) \cdot r(h,j)=0$ holds.\end{proof} 

\label{PS.lemma.minfault}
\end{lemma}
\begin{corollary}
A multi-robot system can tolerate a minor fault of level $\lambda$ if Lemma~\ref{PS.lemma.minfault} holds for the local tasks involved in the fault.
\end{corollary}
\begin{lemma}
A multi-robot system is strongly fault tolerant if and only if for each robot $i$ there exists a set of robots $I$ such that $\mathcal{L}_i \subseteq \cup_{h \in I, h \neq i }^n \mathcal{L}_h$ and the problem \eqref{PS.eq.ass} is consistent under the constraints $r(i,j) \cdot r(h,j)=0$ $\forall j: l_j \in \mathcal{L}_i, h \in I$. 

\begin{proof} (\emph{if}) Since $\mathcal{L}_i \subseteq \cup_{h \in I, h \neq i } \mathcal{L}_h$ all the local task of the robot $i$ can be performed by some other robots in the system. Then when the robot $i$ is asked to perform a local task $l_j$, exists at least another robot that can perform such task, this robot is not deployed since $r(i,j) \cdot r(h,j)=0$ $\forall j: l_j \in \mathcal{L}_i, h \in I$.

 (\emph{only if}) if the system can tolerate a major fault then exists a robot $i$ that is redundant, i.e. its tasks can be performed by other robots. Hence there exists a set of robots $I$, in which $i$ is not contained, such that $\mathcal{L}_i \subseteq \cup_{h \in I, h \neq i } \mathcal{L}_h$, moreover since each local task $l_j$ of the robot $i$ can be performed by a robot $h \in I$, robot $h$ is not deployed when $i$ is performing $l_j$ otherwise it could not take over robot $i$, hence $r(i,j) \cdot r(h,j)=0$ $\forall j: l_j \in \mathcal{L}_i, h \in I$.\end{proof} 
\label{PS.lemma.majfault}
\end{lemma}

\section{Motivating Example}
\label{ME}
\begin{figure}[h]
\centering
\includegraphics[width=0.85\columnwidth]{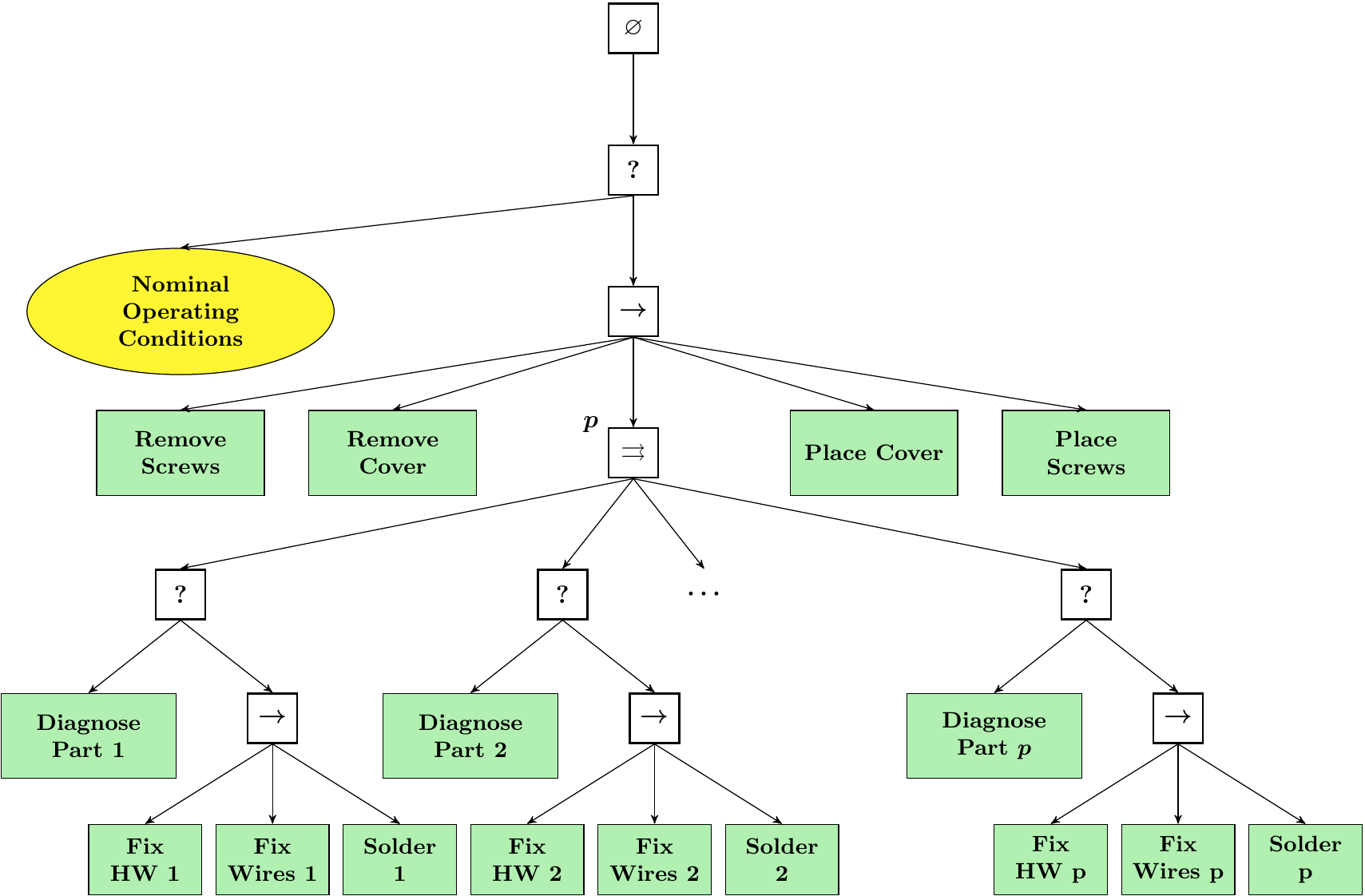}
 \caption{$\bt_G$ of Example~\ref{PS.ex.1}  }\label{PS.fig.tgex}
\end{figure}

In this section we expand upon the example that was briefly mentioned in Section \ref{IN}, and involves
a multi-robot system that is to replace damaged parts of the electronic hardware of a vehicle. Let us assume that to repair the vehicle, the system has to open the metal cover first, then it has to check which parts have been damaged, replace the damaged parts, solder the connecting wires, and finally close the metal cover. In this example the setup is given by a team composed by robots of type $A$, $B$ and $C$. Robots of type $A$ are small mobile dual arm robots that can carry spare hardware pieces, place them and remove the damaged part from the vehicles, and do diagnosis. Robots of type $B$  are also small mobile dual arm robots with a gripper and soldering iron as end effectors. Robots of type $C$ are big dual arm robots that can remove and fasten the vehicle cover. We consider the general case in which the vehicle's electronic hardware consists of $p$ parts, all of them replaceable. The global tasks set is defined as: 
\begin{equation}
\begin{split}
\mathcal{G}=&\{\mbox{Normal Operating Condition}, \mbox{Open Metal Cover},\\ &\mbox{Close Metal Cover},\mbox{Fix Part $1$},\mbox{Fix Part $2$},\ldots,\mbox{Fix Part $p$},\\ &\mbox{Diagnose Part $1$},\mbox{Diagnose Part $2$},\ldots,\mbox{Diagnose Part $p$}\}. \hspace{5cm}
\end{split}
\end{equation}
where:
\begin{equation}
\begin{split}
\mbox{Open Metal Cover}\hspace{0.5mm} = \hspace{0.5mm}&\{\mbox{Remove Screws},\mbox{Remove Cover}\}  \\
\mbox{Close Metal Cover} = \hspace{0.5mm}&\{\mbox{Place Screws},\mbox{Place Cover}\} \\
\mbox{Fix Part $k$}\hspace{4mm} = \hspace{0.5mm}&\{\mbox{Fix HW $k$},\mbox{Fix Wires $k$},  \\ &\mbox{Solder $k$}\} \; k=\{1,2,\ldots,p\}.
\end{split}
\end{equation}
The local tasks for robots of type $A$ set are:
\begin{equation}
\begin{split}
\mathcal{L}_i= \hspace{0.5mm}&\{\mbox{Replace HW}, \mbox{Replace Wires}, \mbox{Do Diagnosis on $k$}\} \hspace{0.5cm}\\ \;&i=\{1,2\}.\hspace{5.5cm}
\end{split}
\end{equation}
The local tasks for robots of type $B$ set are:
\begin{equation}
\begin{split}
\mathcal{L}_i= \hspace{0.5mm}&\{\mbox{Replace HW}, \mbox{Replace Wires},\\ &\mbox{Do Diagnosis on $k$},\mbox{Solder} \}\; i=\{3,4\}.\hspace{5cm}
\end{split} 
\end{equation}
The global tasks for robots of type $C$ set are:
\begin{equation}
\begin{split}
\mathcal{L}_i= \hspace{0.5mm}&\{\mbox{Use Screwdriver}, \mbox{Move Frame} \;i=\{5,6\}. \hspace{5cm}
\end{split}
\end{equation}
The map from global tasks to local tasks $\psi$ are as follows:
\begin{equation}
\begin{split}
&\psi(\mbox{Remove Screws})  \hspace{4.75mm}= \mbox{Use Screwdriver} \\
&\psi(\mbox{Place Screws})  \hspace{8.75mm}= \mbox{Use Screwdriver} \\
&\psi(\mbox{Remove Cover}) \hspace{6.25mm}= \mbox{Move frame} \\
&\psi(\mbox{Place Cover}) \hspace{10.25mm}= \mbox{Move frame} \\
&\psi(\mbox{Diagnose Part $k$}) \hspace{4.25mm} = \mbox{Do Diagnosis on $k$} \\
&\psi(\mbox{Fix HW $k$}) \hspace{12.55mm} = \mbox{Replace HW $k$} \\
&\psi(\mbox{Fix Wires $k$}) \hspace{10.25mm} = \mbox{Replace Wires $k$} \\
&\psi(\mbox{Solder $k$}) \hspace{15.25mm} =  \mbox{Use Soldering Iron on $k$} \\
\end{split}
\end{equation}
The tree $\bt_G$ defined is showed in Fig.~\ref{PS.fig.tgex}. 
\subsection{Execution}
\begin{figure}[h]
\centering
\begin{subfigure}[b]{\columnwidth}
\centering
\includegraphics[width=0.78\columnwidth]{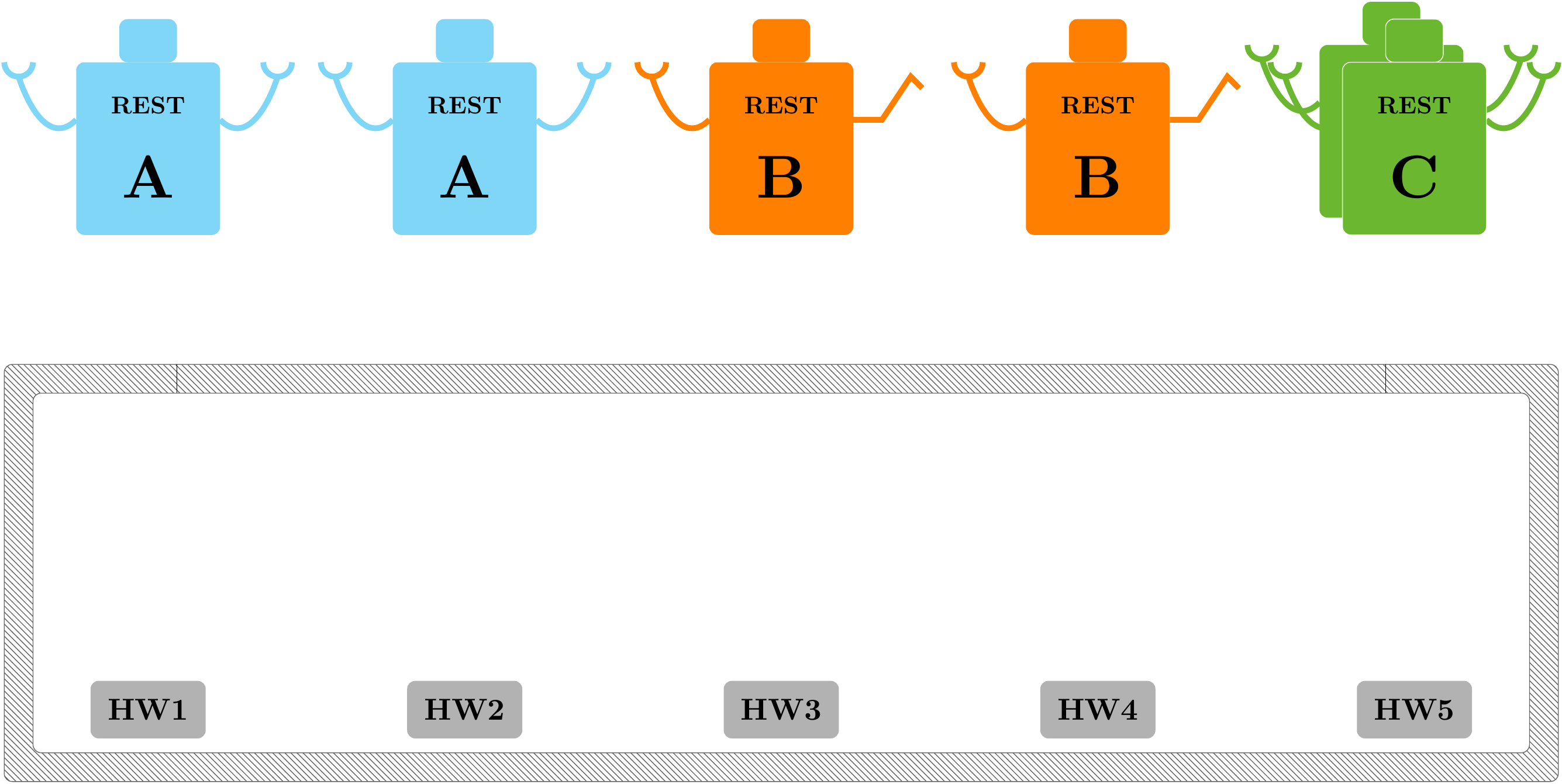}
\caption{Screenshot \rom{1}}
\label{ME.fig.1}
\end{subfigure}
\;
\begin{subfigure}[b]{\columnwidth}
\centering
\includegraphics[width=0.78\columnwidth]{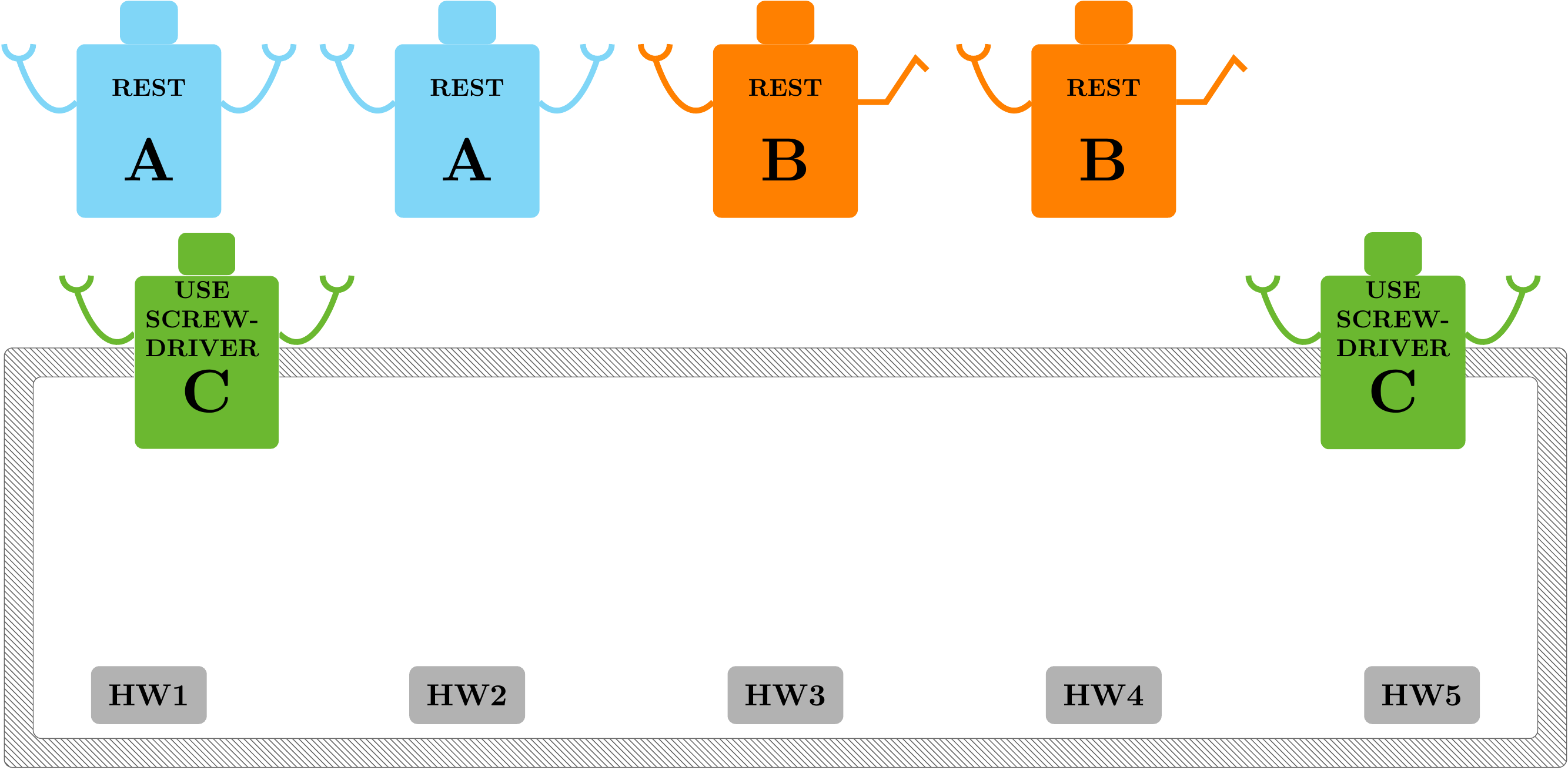}
\caption{Screenshot \rom{2}}
\label{ME.fig.2}
\end{subfigure}
\begin{subfigure}[b]{\columnwidth}
\centering
\includegraphics[width=0.78\columnwidth]{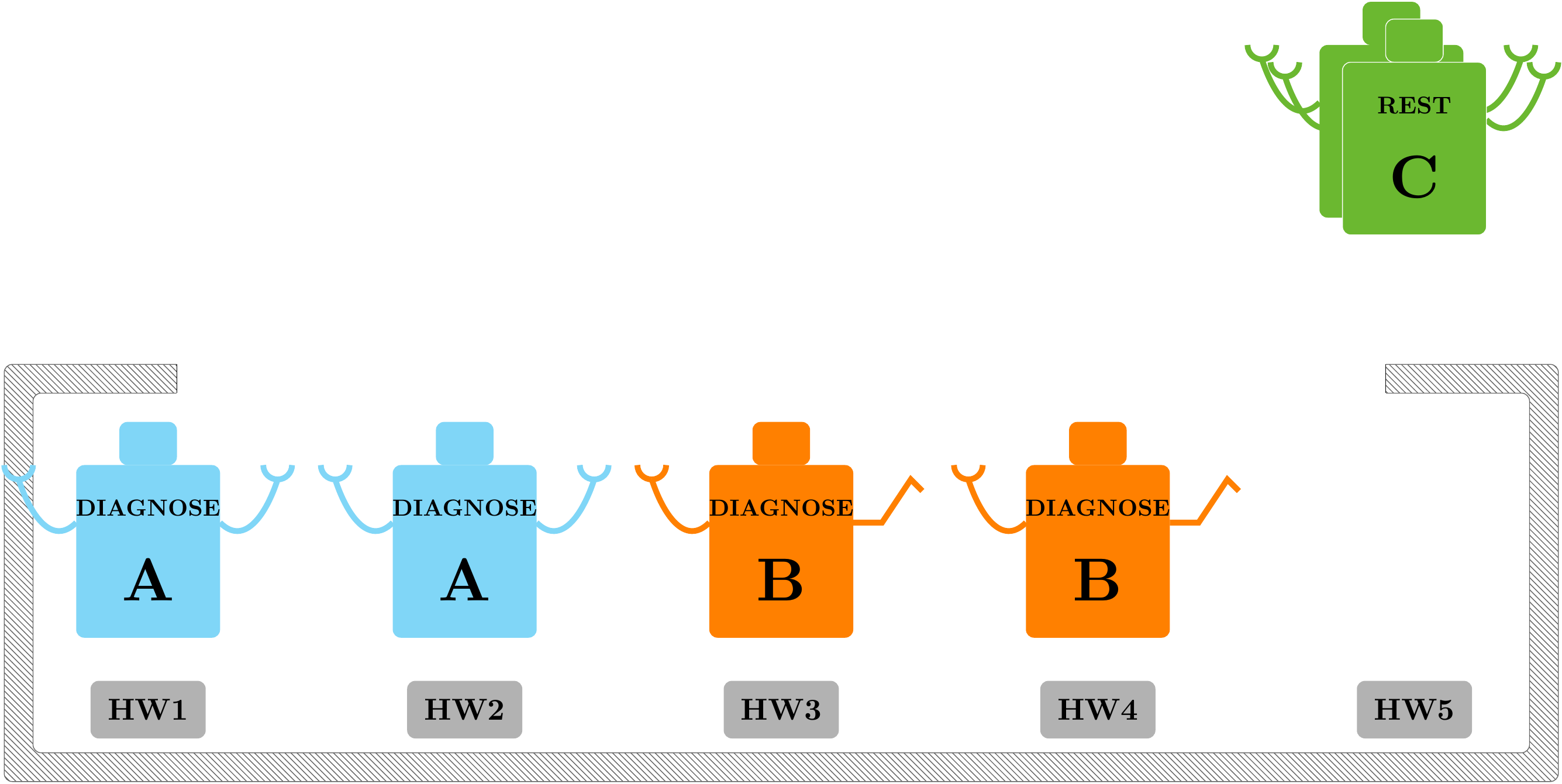}
\caption{Screenshot \rom{3}}
\label{ME.fig.3}
\end{subfigure}
\;
\begin{subfigure}[b]{\columnwidth}
\centering
\includegraphics[width=0.78\columnwidth]{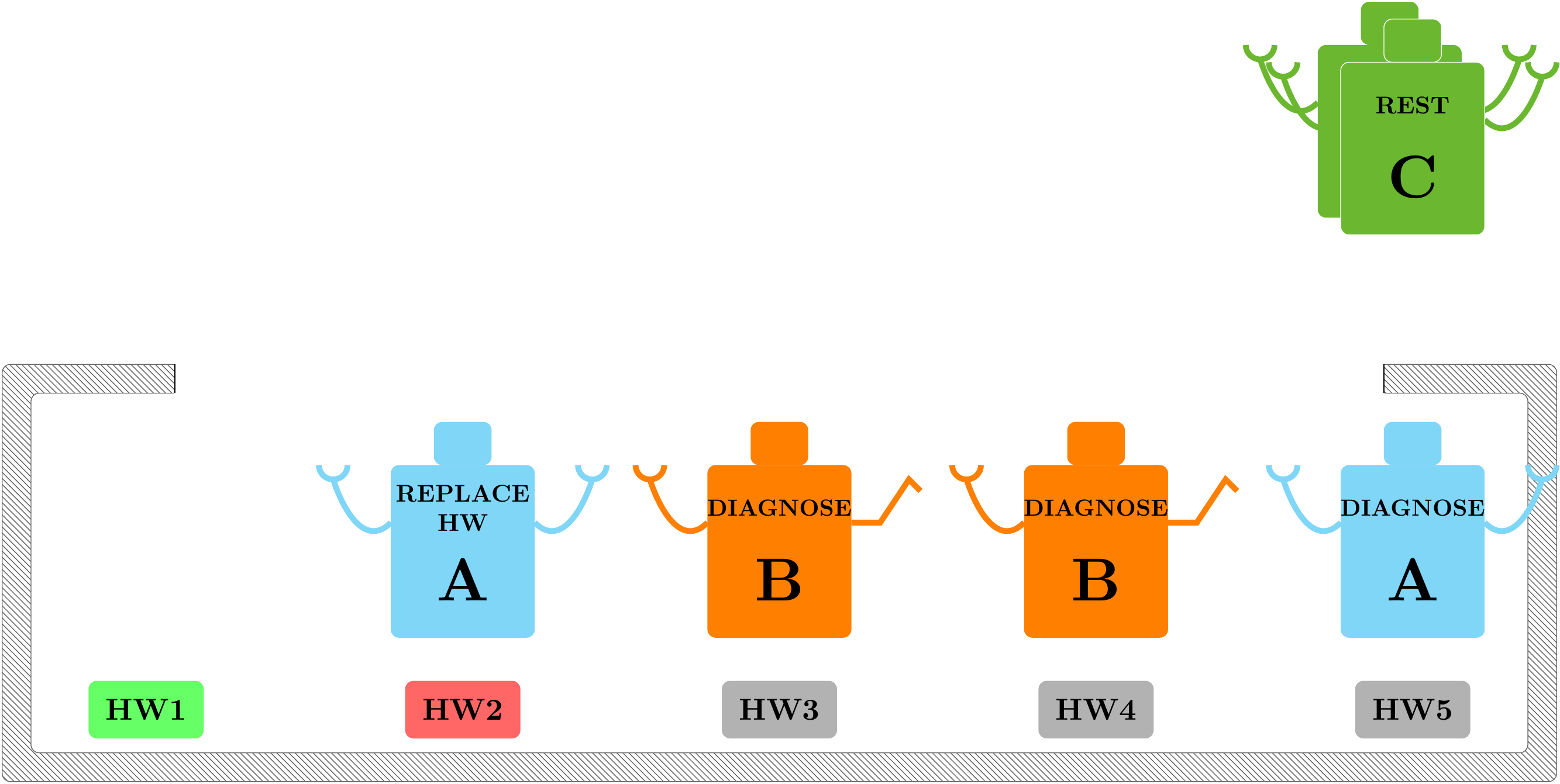}
\caption{Screenshot \rom{4}}
\label{ME.fig.4}
\end{subfigure}
\end{figure}

\begin{figure}[h]

\begin{subfigure}[b]{\columnwidth}
\centering
\includegraphics[width=0.78\columnwidth]{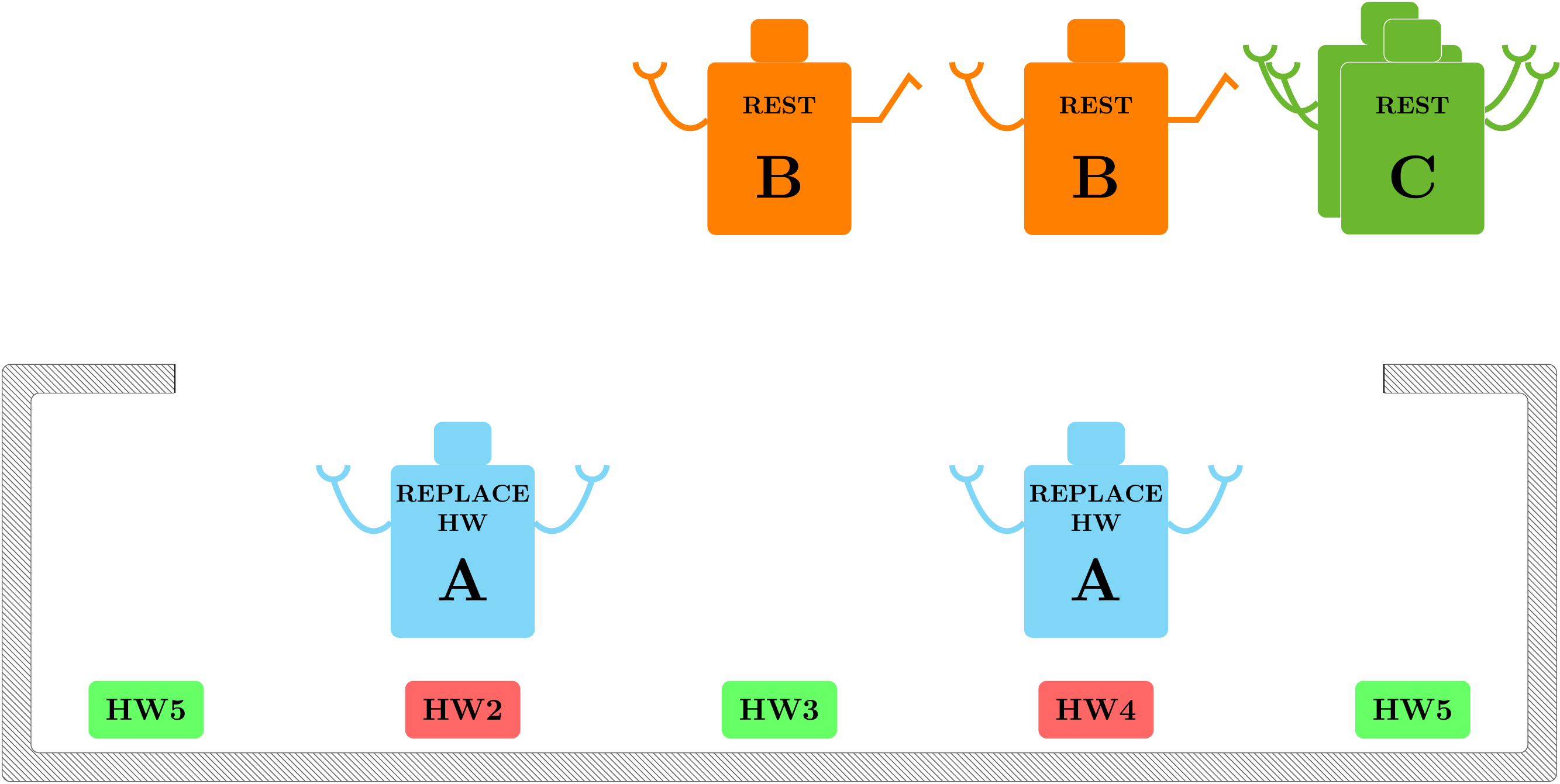}
\caption{Screenshot \rom{5}}
\label{ME.fig.5}
\end{subfigure}
\;
\begin{subfigure}[b]{\columnwidth}
\centering
\includegraphics[width=0.78\columnwidth]{exFig6.pdf}
\caption{Screenshot \rom{6}}
\label{ME.fig.6}
\end{subfigure}
\begin{subfigure}[b]{\columnwidth}
\centering
\includegraphics[width=0.78\columnwidth]{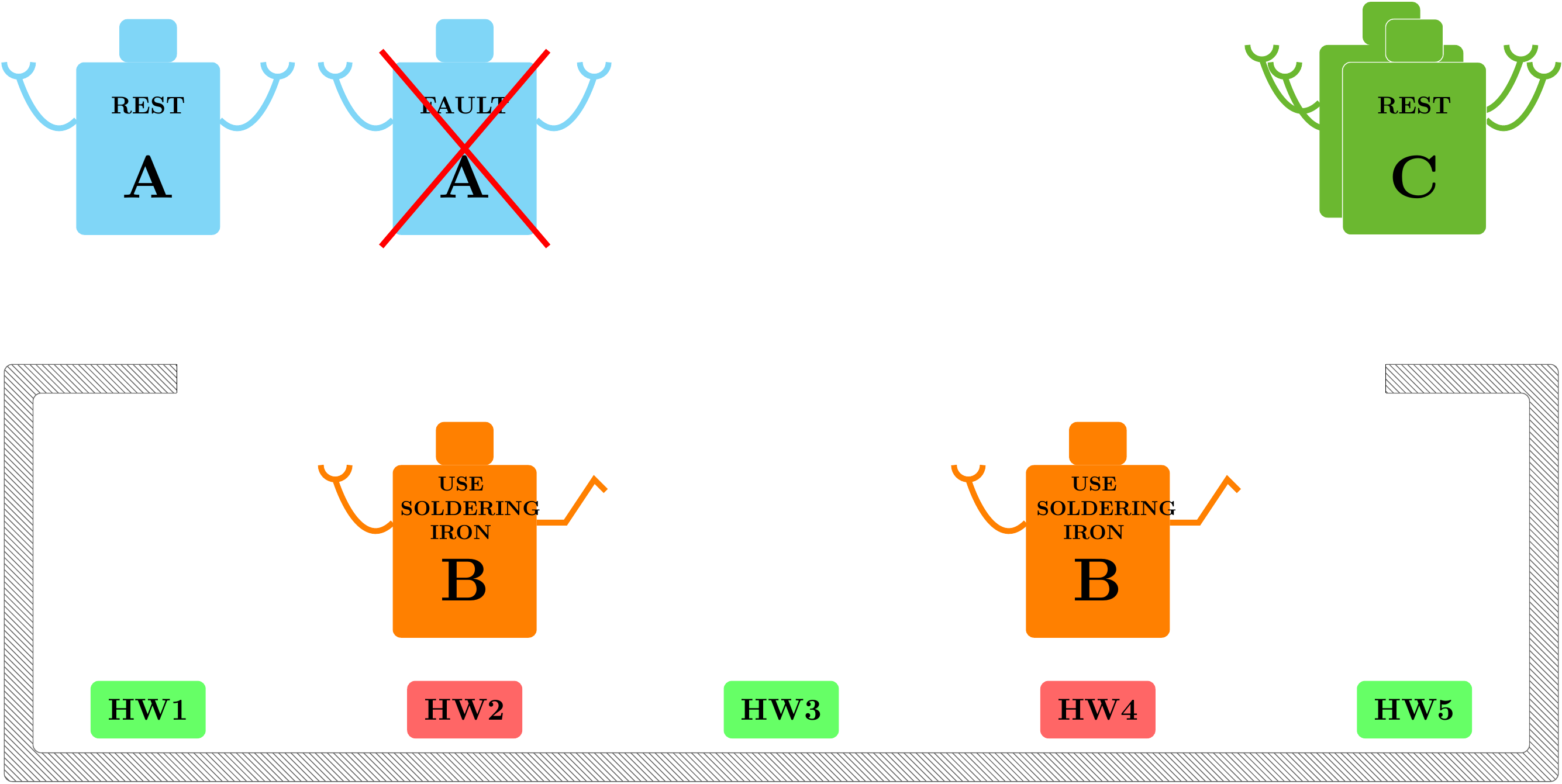}
\caption{Screenshot \rom{7}}
\label{ME.fig.7}
\end{subfigure}
\;
\begin{subfigure}[b]{\columnwidth}
\centering
\includegraphics[width=0.78\columnwidth]{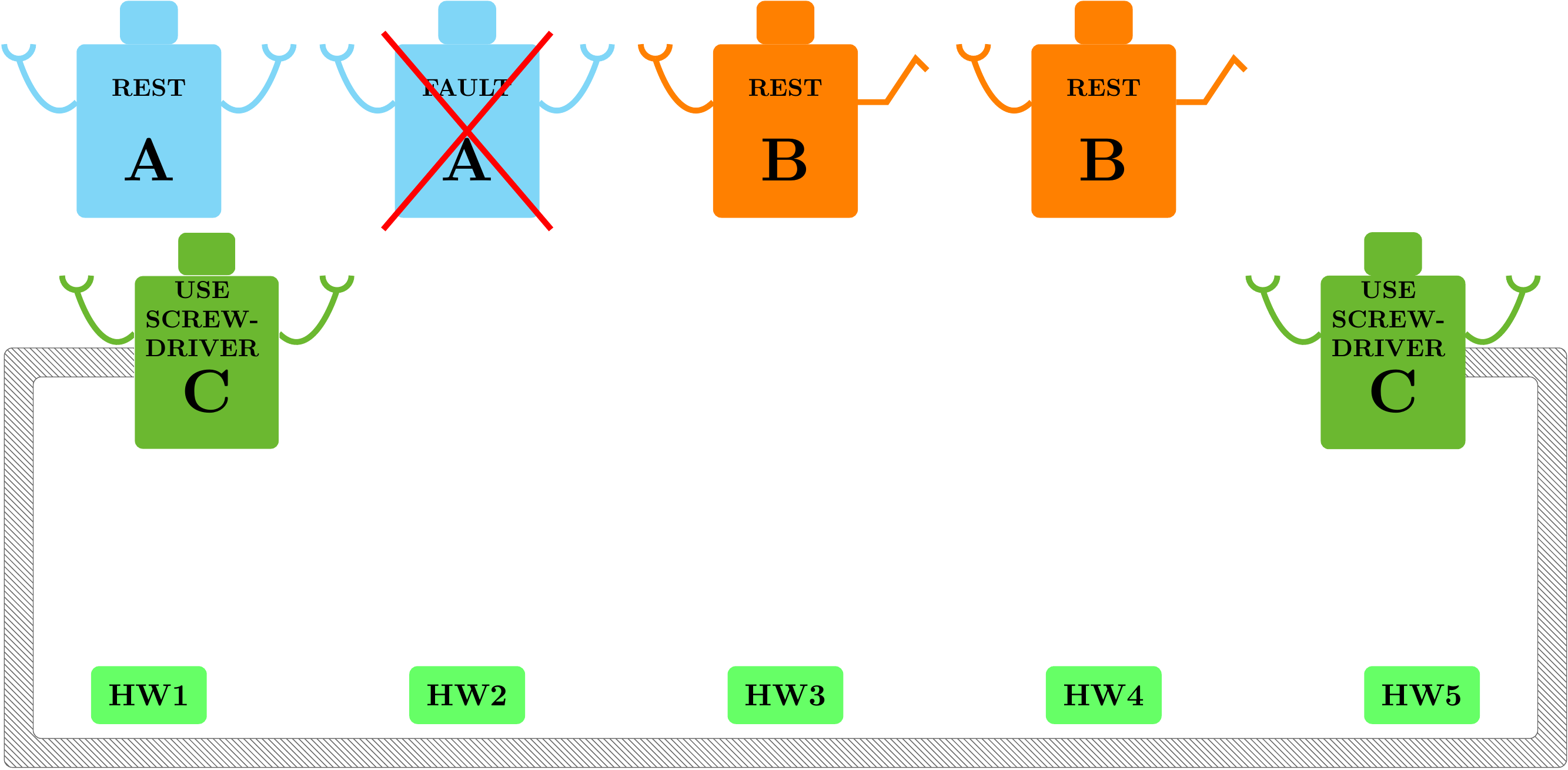}
\caption{Screenshot \rom{8}}
\label{ME.fig.8}
\end{subfigure}
\caption{Plan Execution. HW components are either un-diagnosed (grey), checked and OK (green) or broken (red)}\label{PS.fig.tatg}
\end{figure}

Here we describe the execution of the above mission using the proposed framework.\\ Consider a team of $2$ robots of type $A$, $2$ robots of type $B$, and $2$ robots of type $C$ designate to diagnose and fix a vehicle composed by $5$ critical parts as depicted in Fig.~\ref{ME.fig.1}. The robots' performances $p(i,l_j)$ are related to  how fast tasks are accomplished. They are collected in Table~\ref{PS.tab.per}, and the scenario's parameters are in Table~\ref{PS.tab.par}.\\
\begin{table}[t]
\centering
  \begin{tabular}{| l | c | c | c |}\hline
    Task & Robots $A$ & Robots $B$ & Robots $C$  \\ \hline \hline 
      Use Screwdriver & $\mhyphen \infty$ & $\mhyphen \infty$ & 1 \\ \hline 
      Move Frame & $\mhyphen \infty$ & $\mhyphen \infty$ & 1 \\ \hline 
      Do Diagnosis on $k$ & 3 & 1 & $\mhyphen \infty$ \\ \hline 
      Replace HW $k$ & 1.5 & 1 & $\mhyphen \infty$  \\ \hline 
      Replace Wires $k$ & 1.5 & 1 & $\mhyphen \infty$  \\ \hline 
      Use Soldering Iron on $k$ & $\mhyphen \infty$ & 1 & $\mhyphen \infty$  \\  

               \hline
  \end{tabular}
  \caption{Robot performances values}
  \label{PS.tab.per}	
  \end{table}
\begin{table}[t]
\centering
  \begin{tabular}{| l | c |}\hline
    Patameter & Value  \\ \hline \hline 
      $\nu(\mbox{Remove Screws, Use Screwdriver})$ & 1  \\ \hline 
      $\mu(\mbox{Remove Screws, Use Screwdriver})$ & 2  \\ \hline 
      $\nu(\mbox{Place Screws, Use Screwdriver})$ & 1  \\ \hline 
      $\mu(\mbox{Place Screws, Use Screwdriver})$ & 2  \\ \hline 
      $\nu(\mbox{Remove Cover, Move Frame})$ & 2  \\ \hline 
      $\mu(\mbox{Remove Cover, Move Frame})$ & 2  \\ \hline 
      $\nu(\mbox{Place Screws, Remove Cover})$ & 1  \\ \hline 
      $\mu(\mbox{Place Screws, Remove Cover})$ & 2  \\ \hline 
      $\nu(\mbox{Place Cover, Move Frame})$ & 2  \\ \hline 
      $\mu(\mbox{Place Cover, Move Frame})$ & 2  \\ \hline 
      $\nu(\mbox{Diagnose Part $k$, Do Diagnosis on $k$})$ & 1  \\ \hline 
      $\mu(\mbox{Diagnose Part $k$, Do Diagnosis on $k$})$ & 1  \\ \hline 
      $\nu(\mbox{Fix HW $k$, Replace HW $k$})$ & 1  \\ \hline 
      $\mu(\mbox{Fix HW $k$, Replace HW $k$})$ & 1  \\ \hline 
      $\nu(\mbox{Fix Wires $k$, Replace Wires $k$})$ & 1  \\ \hline 
      $\mu(\mbox{Fix Wires $k$, Replace Wires $k$})$ & 2  \\ \hline 
      $\nu(\mbox{Solder $k$, Use Soldering Iron on $k$})$ & 1  \\ \hline
      $\mu(\mbox{Solder $k$, Use Soldering Iron on $k$})$ & 3  \\
               \hline
  \end{tabular}
  \caption{Scenario parameters. Note that when $\nu=\mu$, as in the case of (Remove Cover, Move Frame) we have that the number of required robots for a task is equal to the maximal number of robots that can participate in executing the task. 
  }
  \label{PS.tab.par}	
  \end{table}  
When the vehicle is running on nominal operating conditions, the assignment tree $\bt_A$ of each robot has $a_j=b_j=0$ since no assignments are needed. When a fault on the vehicle takes place, the condition ``Nominal Operating Conditions" is no longer true in all the robots' BTs, then the ``Remove Screws" action has to be executed. According to $\psi(\mbox{Remove Screws})$ this global task requires the local task ``Use Screwdriver" to be accomplished. The constraints related to this task have to be added in~\eqref{PS.eq.ass} after checking the feasibility of the optimization problem. The parameter $a_j$ and $b_j$ for ``Use Screwdriver" are changed into: $a_j=\nu(\mbox{Remove Screws, Use Screwdriver})=1$ and $b_j=\mu(\mbox{Remove Screws, Use Screwdriver})=2$ with $j: l_j=\{\mbox{Use Screwdriver}\}$. Solving~\eqref{PS.eq.ass} both robots $C$ are assigned to ``Use Screwdriver", while the other are not deployed (Fig.~\ref{ME.fig.2}). When a robot accomplishes its local task, then the corresponding $b_j$ is decreased by $1$ until $b_j=0$ and the robot can be deployed for a new task if needed. A similar assignment is then made to accomplish the global task ``Remove Cover". After removing the metal cover, all the vehicle hardware have to be diagnosed, and the global task ``Diagnose Part 1" will change the parameters in $a_j=\nu(\mbox{Diagnose Part 1, Do Diagnosis on 1})=1$ and $b_j=\mu(\mbox{Diagnose Part 1, Do Diagnosis on 1})=1$ with $j:l_j=\mbox{Do Diagnosis on 1}$, in a similar way ``Diagnose Part 2" will change the parameters in $a_j=\nu(\mbox{Diagnose Part 2, Do Diagnosis})=1$ and $b_j=\mu(\mbox{Diagnose Part 2, Do Diagnosis})=1$ with $j:l_j=\mbox{Do Diagnosis}$ and so forth until ``Diagnose Part 4". When the system changes the parameters related to ``Diagnose Part 5" the optimization problem is not feasible since there are not enough agents, then the assignment for ``Diagnose Part 5" does not take place (Fig.~\ref{ME.fig.3}).\\
Here we show the reactive functionality of the proposed framework. Note that the robots of type $A$  have higher performance than the robots of type $B$ in doing diagnosis according to Table~\ref{PS.tab.per}, i.e., they are faster. While the robots of type $B$ are still doing diagnosis on $HW3$ and $HW4$, the robots of type $A$ can be deployed for new tasks without waiting for the slower robots (Fig.~\ref{ME.fig.4}). After diagnosis, $HW2$ and $HW4$ must be replaced, robots of type $A$ have higher performance in ``Replace HW" therefore, they are deployed to minimize the objective function of~\eqref{PS.eq.ass}, robots  of type $B$ are not deployed since $\mu(\mbox{Fix HW $k$, Replace HW $k$})	= 1$ i.e. the replacing of an electronic HW is made by a single robot (Fig.~\ref{ME.fig.5}).\\ If a robot of type $A$ has a major fault, all its performance parameters are set to $\mhyphen \infty$ and a robot of type $B$ can take over since it can execute all the task of a robot $A$, this situation is depicted in Fig.~\ref{ME.fig.6}. Here  two robots of type $B$ are replacing wires on $HW1$, since $\mu(\mbox{Fix Part i, Replace Wires})=2$ and a robot of type $A$ is replacing wires on $HW2$. The final part of the repair is the soldering of the new wires, executable only by robots of type $B$ (Fig.~\ref{ME.fig.7}). Finally the metal cover is put back (Fig.~\ref{ME.fig.8})  and the tree $\bt_G$ returns success.
\label{PS.ex.1}

\subsection{Fault Tolerance Analysis}
Analyzing the system, it is neither weakly nor strongly fault tolerant. A minor fault on the local task ``Move Frame" can not be tolerated since  $\nu(\mbox{Remove Cover, Move Frame})=2$ as well as  $\nu(\mbox{Place Cover, Move Frame})=2$ thus a single robot cannot perform the task. However, some faults can be tolerated. The multi-robot system can tolerate up to $3$ major faults ($2$ faults involving both robots $A$, and $1$ fault involving a robot $B$) and up to $11$ minor faults ($1$ fault involving ``Use Screwdriver", $3$ faults involving ``Replace HW $k$", $3$ faults involving ``Replace Wires $k$", and $3$ faults involving ``Do diagnosis on $k$" ), as long as at least one robot of type $B$ and $2$ robots of type $C$ are operating. Note that the local task ``Use soldering iron on $k$" can be performed by a robot of type $B$ only.
\section{Conclusions}
\label{Conc}
We have shown a possible use of BTs as a distributed controller for multi-robot systems working towards global goals. This extends the fields where BTs can be used, with a strong application in robotic systems. We have also shown how using BTs as a framework for robot control becomes natural going from a plan meant to be executed by a single robot to a multi-robot plan.
\section*{Acknowledgment}
This work has been supported by the Swedish Research Council (VR) and the European Union Project RECONFIG, (FP7-ICT-2011-9), the authors gratefully acknowledge the support.
\bibliographystyle{IEEEtran}
\bibliography{references,hybrid,behaviorTreeRefs}
\end{document}